\documentclass[letterpaper, 10 pt, conference]{ieeeconf}  

\IEEEoverridecommandlockouts                              

\usepackage{graphics} 
\usepackage{epsfig} 
\usepackage[table]{xcolor}
\definecolor{grey}{rgb}{0.5,0.5,0.5}
\usepackage{booktabs}
\usepackage{subcaption}
\usepackage{hyperref}
\usepackage{amsmath}
\usepackage{algorithm}
\usepackage{algpseudocode}
\usepackage{float}
\usepackage{url}
\usepackage{caption}
\usepackage{amssymb}  

\usepackage[
    style=ieee,
    natbib=true,
    citestyle=numeric-comp,
    doi=false,
    isbn=false,
    url=false]{biblatex} 

\title{\LARGE \bf
UniTracker: Learning Universal Whole-Body Motion Tracker for Humanoid Robots
}

\author{Kangning Yin$^{1,2,3*}$, Weishuai Zeng$^{2,4*}$, Ke Fan$^{1, 2}$, Minyue Dai$^{2, 6}$ Zirui Wang$^{2,3,5}$,Qiang Zhang$^{7}$, \\ Zheng Tian$^{8}$, Jingbo Wang$^{2}$, Jiangmiao Pang$^{2}$, Weinan Zhang$^{1,2,3}$ \\
\\
$^{1}$Shanghai Jiao Tong Univeristy, $^{2}$Shanghai Artificial Intelligence Laboratory, $^{3}$Shanghai Innovation Institute, \\$^{4}$Peking University, $^{5}$Zhejiang University, $^{6}$Fudan University \\
$^{7}$The Hong Kong University of Science and Technology (Guangzhou), $^{8}$ShanghaiTech University\\
\\
Paper Website: 
\href{https://yinkangning0124.github.io/Humanoid-UniTracker/}{https://yinkangning0124.github.io/Humanoid-UniTracker/}}

\begin{document}

\maketitle

\thispagestyle{empty}
\pagestyle{empty}

\begin{abstract}

Achieving generalizable whole-body motion control is essential for deploying humanoid robots in real-world environments. However, existing MLP-based policies trained under partial observations often suffer from limited expressiveness and struggle to maintain global consistency. These shortcomings manifest as less expressive motion, orientation drift, and poor generalization across diverse behaviors.
To address these limitations, we propose UniTracker, a three-stage framework for scalable and adaptive motion tracking. The first stage learns a privileged teacher policy that produces high-fidelity reference actions. Building on this, the second stage trains a CVAE-based universal policy that captures a global latent representation of motion, enabling robust performance under partial observations. Crucially, we align the partial-observation prior with a full-observation encoder, injecting global intent into the latent space. In the final stage, a lightweight adaptation module fine-tunes the student policy on challenging sequences, supporting both per-instance and batch adaptation.
We validate UniTracker in simulation and on a Unitree G1 humanoid robot, demonstrating superior tracking accuracy, motion diversity, and deployment robustness compared to current baselines.

\end{abstract}

%
\section{INTRODUCTION}
Humanoid robots have garnered growing interest in the robotics community for their human-like morphology, which equips them with the potential to perform a wide range of tasks traditionally carried out by humans. To function effectively in real-world, human-centric environments, these robots must exhibit not only physical versatility but also robust and expressive motor control. Among the key enablers of such capabilities is whole-body control, which coordinates multiple joints and limbs to perform complex tasks while ensuring stability, expressiveness and adaptability.

Recent research has explored various control interfaces tailored to humanoid whole-body controller. These can be broadly categorized into dense and sparse control signals. Dense signals, such as teleoperation~\citep{ze2025twistteleoperatedwholebodyimitation, lu2025mobiletelevisionpredictivemotionpriors, fu2024humanplushumanoidshadowingimitation,  he2024learninghumantohumanoidrealtimewholebody, qiu2025humanoidpolicyhuman, he2024omnih2ouniversaldexteroushumantohumanoid}, offline motion datasets~\citep{shao2025langwbclanguagedirectedhumanoidwholebody, shi2025adversariallocomotionmotionimitation, he2025asapaligningsimulationrealworld, ji2025exbody2advancedexpressivehumanoid, cheng2024expressivewholebodycontrolhumanoid, chen2025gmt, xie2025kungfubot}, and video-based motion estimation~\citep{mao2024learningmassivehumanvideos, allshire2025visualimitationenablescontextual}, provide rich trajectory-level information. In contrast, sparse signals such as high-level task commands and VR-based guidance~\citep{he2025hoverversatileneuralwholebody, xue2025unifiedgeneralhumanoidwholebody} offer minimal information and often lead to reduced motion quality. In this work, we focus on \textbf{universal whole-body motion tracking}, where the input is reference motion sequences and the goal is to track them robustly and expressively using a single policy.

\begin{figure}
    \centering
    \vspace{-2mm}
    \includegraphics[width=\columnwidth]{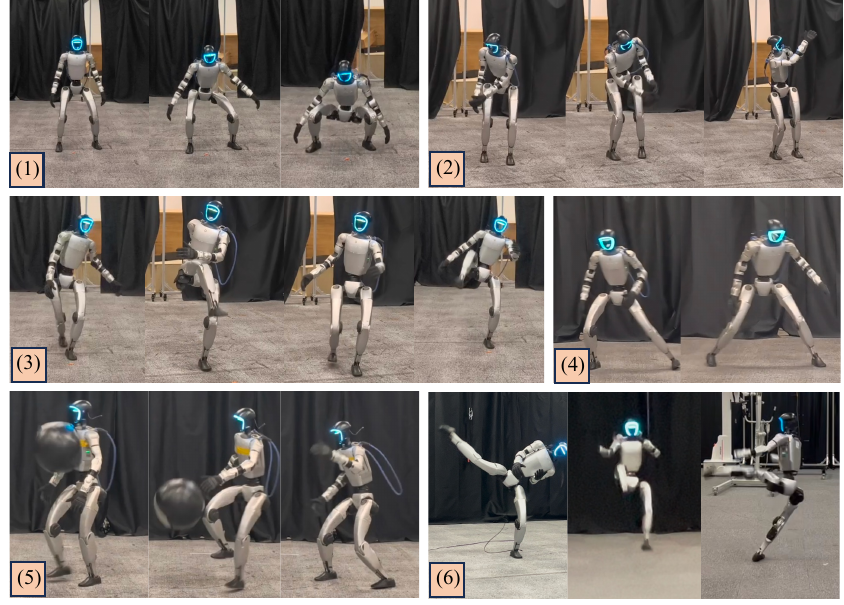}
    \caption{We deploy our UniTracker on a real humanoid robot, enabling it to perform a diverse range of motions, including (1)squat, (2)golf, (3)high kick, (4)ateral step, (5)dance under external force and (6) challenging motions by fast adaption.}
    \vspace{-6mm}
    \label{fig:teaser}
\end{figure}

A widely adopted paradigm for learning motion tracking policies is the teacher-student framework. In this paradigm, a teacher policy is initially trained using full privileged observations to precisely track reference motions within simulation environments. Subsequently, a student policy is learned to imitate the teacher policy based solely on the partial observations available during deployment. Despite its effectiveness, existing implementations of this framework, particularly those utilizing simple MLP-based DAgger architectures~\citep{ross2011reductionimitationlearningstructured} encounter three significant limitations. First, they frequently fail to preserve the diversity inherent in the original reference motions during the distillation process, leading to behavior that is averaged and less expressive. Second, constrained by their limited representational capacity, such models often exhibit poor generalization to unseen motion sequences. Third, the absence of global contextual information during training gives rise to issues such as orientation drift and broader inconsistencies in global behavior, which become particularly pronounced when the policies are deployed on real-world humanoid robots.

To address the aforementioned limitations of existing teacher-student frameworks, we introduce UniTracker, a unified and expressive whole-body tracking framework that integrates a Conditional Variational Autoencoder (CVAE)~\citep{sohn2015learning} into the student policy architecture. By explicitly modeling a structured latent space conditioned on future motion references, UniTracker enables the policy to generate diverse and high-fidelity behaviors even under partial observations.
From a probabilistic standpoint, the latent variable captures the inherent ambiguity in the mapping from observations to actions, allowing the policy to model a distribution over plausible motor behaviors rather than collapsing to a single deterministic output. This capability enhances motion expressiveness and significantly improves generalization across diverse and unseen motion patterns.

In addition, the CVAE-based framework also effectively addresses the challenge of missing global context—often manifested as orientation drift and other global inconsistencies during deployment. To this end, we employ task-aware feature modeling during training: the encoder is trained using privileged, globally informative observations to infer a structured latent representation, while a prior network is concurrently trained based only on the partial observations available at deployment time. The two distributions are aligned via a KL divergence objective. As a result, although the final deployed policy operates under partial observability, it benefits from a latent space informed by global context during training. This implicit incorporation of global information leads to more coherent and globally consistent behaviors in real-world settings.

While this CVAE-based universal policy demonstrates strong performance across a wide range of motions, it is neither necessary nor realistic to expect it to perfectly track all possible reference sequences—particularly those that are rare, highly dynamic, or lie far outside the training distribution. To accommodate such challenging cases, we introduce a fast adaptation phase that fine-tunes the universal policy in a task-specific manner. Leveraging the expressiveness and generality of the base policy, this adaptation process enables rapid specialization with minimal training time. Moreover, our framework supports both single-sequence adaptation and batch-mode adaptation, allowing for scalable refinement when dealing with multiple difficult motions. This final phase complements the universal policy by extending its practical applicability to edge cases, and highlights the modularity and flexibility of our overall three-stage training framework.

We extensively evaluate UniTracker in both simulated environments and real-world deployment scenarios. Experiments conducted on a 29-DoF Unitree G1 humanoid demonstrate that our policy is capable of tracking over 8,100 diverse motion sequences—including highly dynamic behaviors—using a single unified network. Compared to strong teacher-student baselines that do not incorporate CVAE-based modeling, UniTracker consistently achieves superior performance in terms of tracking accuracy, robustness, and generalization to unseen motions.

\begin{figure*}
    \centering
    \vspace{-2mm}
    \includegraphics[width=\textwidth]{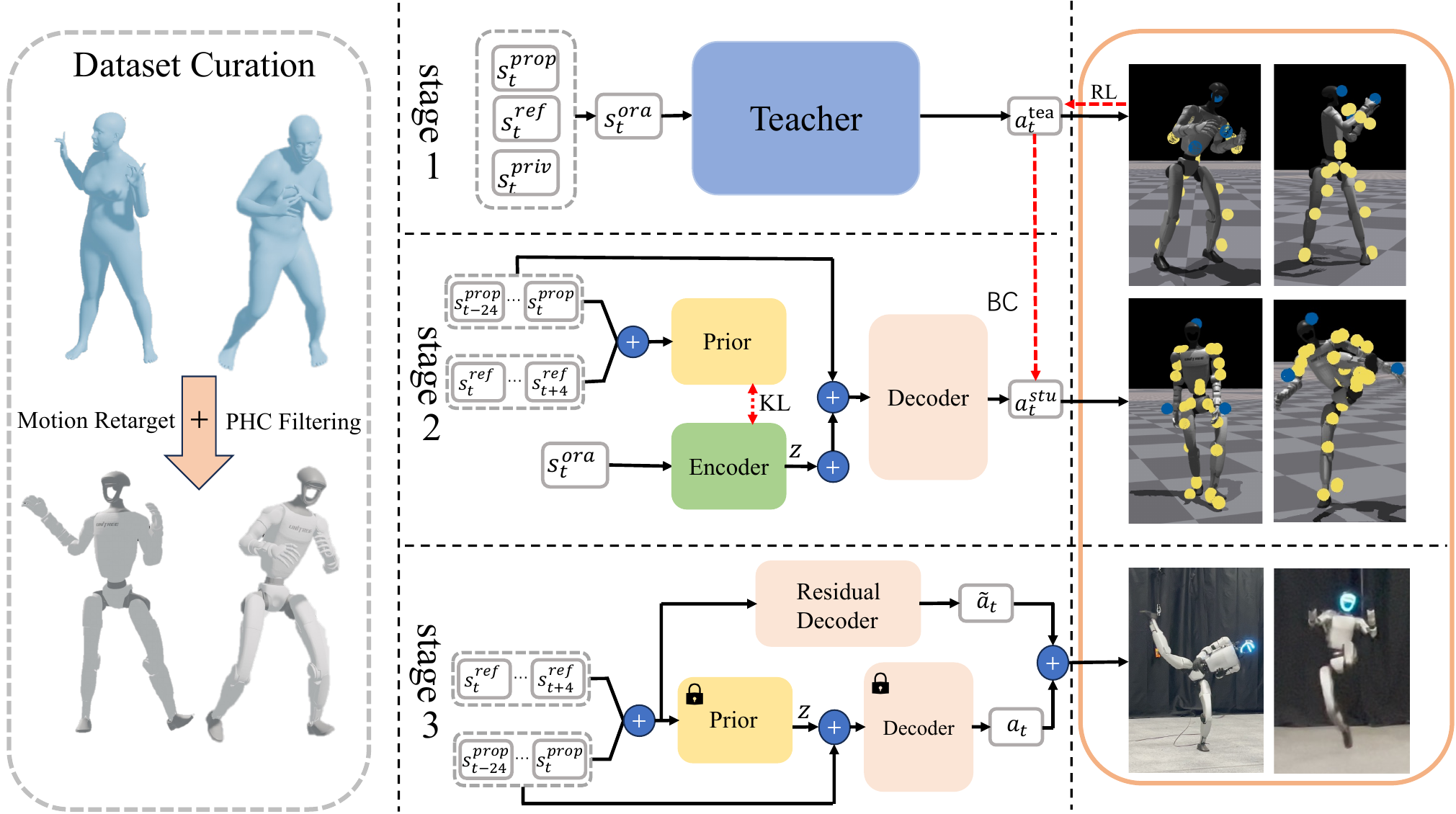}
    \caption{\textbf{An overview of UniTracker}: In Stage 1, we train a teacher policy using oracle states via goal-conditioned reinforcement learning. In Stage 2, we distill the policy into a deployable form using a CVAE-based DAgger framework. In Stage 3, we introduce a fast adaptation module for handling challenging motion sequences, implemented using a residual decoder. The training dataset is derived from the AMASS dataset, filtered by PHC to remove physically infeasible motions.}
    \label{fig:pipeline}
\end{figure*}

Our main contributions are summarized as follows: i)~\textbf{A three-stage training framework for universal whole-body tracking: }  We design a modular pipeline comprising a privileged teacher for data generation, a CVAE-based student policy for deployment under partial observations, and a lightweight fast adaptation phase for motion-specific fine-tuning. ii)~\textbf{Diversity-aware and global context-integrated policy via CVAE modeling:} We employ a Conditional Variational Autoencoder to capture motion diversity and encode global context, enabling expressive behaviors and reducing global inconsistencies by aligning a globally-informed encoder with a partial-observation prior. iii)~\textbf{Fast adaptation on top of a universal policy:} We introduce a rapid fine-tuning mechanism that adapts the universal policy to challenging or out-of-distribution motions, supporting both single-sequence and batch-level adaptation with minimal overhead. iv)~\textbf{Extensive validation on real-world hardware:} We demonstrate that UniTracker robustly tracks over 8,100 human motions on a 29-DoF humanoid using a single policy, outperforming strong teacher-student baselines in accuracy, robustness, and generalization.

\section{METHOD}
\subsection{Problem Formulation}
We formulate the problem of humanoid robot whole-body motion tracking as a goal-conditioned reinforcement learning (RL) task, where a policy $\pi$ is trained to track reference motions at the whole-body level. The state $s_t$ comprises the robot's proprioceptive information $s_t^p$ and the goal $s_t^g$ which specifies the target state for all body parts. The reward function $r_t = R(s_t^p, s_t^g)$, defined in terms of the agent’s proprioception and goal state, yields dense signals to guide policy optimization. To better focus on motion tracking at the whole-body level, we fix the wrist joints of our 29-degree-of-freedom (DoF) Unitree G1 robot\citep{unitreeg1}, reducing the action space to 23 dimensions. The action $a_t \in \mathbb{R}^{23}$ specifies target joint positions, which are executed via a PD controller to actuate the robot. For policy optimization, we employ Proximal Policy Optimization (PPO)~\citep{schulman2017proximalpolicyoptimizationalgorithms} to maximize the expected cumulative discounted reward $\mathbb{E}\left[\sum_{t=1}^T \gamma^{t-1} r_t\right]$.



The remainder of this section is organized as follows. Section~\ref{sec:B} introduces the construction of a high-quality humanoid motion dataset for policy training. Section~\ref{sec:C} describes the training of an oracle policy in simulation, which is designed both to maximize the expressiveness of motion tracking. Section~\ref{sec:D} details the distillation of the oracle policy into a deployable student using a CVAE-based framework. Section~\ref{sec:E} presents our fast adaptation strategy, which enables rapid fine-tuning based on the universal tracker. An overview of the proposed universal whole-body motion tracking framework is provided in Figure~\ref{fig:pipeline}.

\subsection{Humanoid Motion Dataset Curation}
\label{sec:B}
A large-scale humanoid motion dataset serves as fuel for training a universal motion tracker. Our dataset is primarily derived from the publicly available AMASS dataset~\cite{mahmood2019amassarchivemotioncapture}. We first exclude sequences involving human-object interactions and discard clips shorter than 10 frames. We then apply the filtering protocol introduced in PHC~\cite{luo2023perpetualhumanoidcontrolrealtime} to further refine the dataset, ensuring high-quality motion sequences suitable for whole-body tracking. This yields a final training set of 8179 human motions, represented using the SMPL~\cite{SMPL:2015} parameters. The SMPL model parameterizes the human body through shape parameters $\beta \in \mathbb{R}^{10}$, pose parameters $\theta \in \mathbb{R}^{24\times 3}$ and root translation $p\in \mathbb{R}^{3}$. $S$ denotes the SMPL function, where $S(\beta, \theta, p): \beta, \theta,p \rightarrow \mathbb{R}^{6980\times 3}$ maps the SMPL parameters to the positions of vertices of a triangular human mesh.

To bridge the embodiment gap between SMPL human model and humanoid robots, we employ a two-stage retargeting approach inspired by H2O~\cite{he2024learninghumantohumanoidrealtimewholebody}. First, we carefully select 16 corresponding body links and optimize the shape parameter $\beta'$ for humanoid robots by minimizing the distances between selected links in the rest pose. Second, leveraging the optimized $\beta'$ alongside the original pose $\theta$ and translation $p$ from dataset, we perform gradient descent over the humanoid robot's root translation, root orientation and joint positions to minimize the distances between selected links throughout the whole sequence. Additional regularization terms are added to avoid aggressive behaviors and ensure temporal smoothness.

\subsection{Oracle Policy Training in Simulation}
\label{sec:C}
\noindent \textbf{Oracle State Space Design.} We train an oracle motion tracking policy $\pi^{orcale}(a_t|s_t^{p-oracle}, s_t^{g-oracle})$ with all the state information accessible in simulators. The proprioception is defined as $s_t^{p-oracle}\triangleq [p_t, q_t, \theta_t, \dot{p_t}, \dot{q_t}, \omega_t, a_{t-1}]$, which contains the humanoid rigid-body position $p_t$, orientation $\theta_t$, linear velocity $\dot{p_t}$, angular velocity $\omega_t$, joint position $q_t$, joint velocity $\dot{q_t}$ and previous action $a_{t-1}$. The goal state is defined as $s_t^{g-oracle}\triangleq [\hat{p}_{t+1}-p_t, \hat{q}_{t+1}-q_t, \hat{\theta}_{t+1}\ominus\theta_t, \hat{v}_{t+1}-v_t, \hat{\omega}_{t+1}-\omega_t, \hat{p}_{t+1}-p_t^{root}, \hat{\theta}_{t+1} \ominus \theta_t^{root}]$, which contains the one-frame difference between the reference pose $(\hat{p}_{t+1}, \hat{q}_{t+1}, \hat{\theta}_{t+1}, \hat{v}_{t+1}, \hat{\omega}_{t+1})$ and the current pose. $p_t^{root}$ refers to the root translation and $\theta_t^{root}$ refers to the root orientation of the current pose. All these states are rotated to the local coordinate of the current pose.

\noindent \textbf{Reward Design.}
We formulate the reward $r_t$ as a weighted sum of three components: 1) task rewards for motion tracking, 2)regularization, and 3) penalty, detailed in Table~\ref{table:reward function}. We apply curriculum learning to the regularization terms and penalty terms such that the policy could better focus on the motion tracking task itself and gradually take the penalty and regularization into account for more reasonable behaviors. 

\noindent \textbf{Early Termination and Reference State Initialization.}
In the early stage of training, the agent is prone to falling, resulting in the collection of invalid data that hinders effective learning. To address this issue, we follow prior works~\cite{Peng_2018, luo2023perpetualhumanoidcontrolrealtime} and introduce two early termination conditions: 1) orientation: the projected gravity on x or y axis exceeds 0.8; 2) tracking tolerance: the average link distance between the robot and reference motions is further than 0.5m. Proper task initialization is also crucial for RL training. We employ the Reference State Initialization~\cite{Peng_2018, he2025asapaligningsimulationrealworld} framework, where the starting point of the reference motion is randomly sampled for the policy to track. The robot's initial state, including the root position, orientation, linear and angular velocities, as well as joint positions and velocities, is then derived from the corresponding reference pose. This initialization strategy substantially enhances motion tracking training by enabling the policy to learn different motions phases in parallel, rather being constrained to a sequential learning process.

\noindent \textbf{Domain Randomization.}
Domain randomization is a key technique for improving sim-to-real robustness~\cite{cheng2024expressivewholebodycontrolhumanoid, he2024learninghumantohumanoidrealtimewholebody, he2024omnih2ouniversaldexteroushumantohumanoid}. Existing strategies generally fall into two categories: asset property randomization and environmental dynamics randomization. The former modifies physical parameters such as friction, center of mass, and link mass to prevent overfitting to a specific asset configuration. The latter perturbs dynamic elements such as PD gains, torque noise, or external forces to simulate real-world uncertainty.
While prior work often applies both types concurrently, we decouple them to better balance tracking expressiveness and robustness. In the first stage, we randomize only asset properties, allowing the teacher policy to focus on high-fidelity motion tracking. This design provides a strong performance upper bound and serves as a proxy agent for generating high-quality reference actions.
\subsection{Hierarchical Controller via Online Distillation}
\label{sec:D}
\noindent \textbf{Deployable State Space Design.} 
\begin{table}
    \centering
    \small
    \setlength{\tabcolsep}{1.0pt}
    \renewcommand{\arraystretch}{0.5}
    \begin{tabular}{>{\centering\arraybackslash}p{3.0cm} >{\centering\arraybackslash}p{3.0cm} >{\centering\arraybackslash}p{2.0cm}}
        \toprule
        \multicolumn{1}{c}{\textbf{Name}}& \multicolumn{1}{c}{\textbf{Function}} & \multicolumn{1}{c}{\textbf{Weight}}\\
        \midrule
        tracking keypoints  &  $exp(||{p_t}^{ref}-p_t||)$ & 1.6 \\
        \midrule
        tracking feet position & $exp(||{p_{t_{feet}}}^{ref}-p_{t_{feet}}||)$ & 2.1 \\
        \midrule
        tracking body rotation & $exp(||r_t - {r_t}^{ref}||)$ & 0.5 \\
        \midrule
        tracking joint position & $exp(||{q_t}^{ref}-q_t||)$ & 0.75\\
         \midrule
        tracking joint velocity& $exp(||\dot{{q_t}^{ref}}-\dot{q_t}||)$ & 0.5\\
        \midrule
        tracking body linear velocity & $exp(||\dot{{p_t}^{ref}}-\dot{p_t}||)$ & 0.5\\
        \midrule
        tracking body angular velocity & $exp(||\dot{r_t} - \dot{{r_t}^{ref}||})$ & 0.5 \\
        \midrule
        action rate & $-||a_t - a_{t-1}||$ & -0.5\\
        \midrule
        torque & $-\tau$ & -1e-6\\
        \midrule
        slippage & $-||{v_t}^{foot}*{F_t}^{contact}||$ & -1.0\\
        \midrule
        termination &  1.0 & -200.0 \\
    \bottomrule
    \end{tabular}
    \caption{Definition of Reward Functions}
    \vspace{-6mm}
    \label{table:reward function}
\end{table}
Since certain privileged information in the oracle state space is unavailable in real-world deployment, we define a deployable state space based on data accessible on humanoid robots. The proprioception is defined as $s_t^{p-deploy}\triangleq [q_{t-25:t}, \dot{q}_{t-25:t}, w_{t-25:t}^{root}, g_{t-25:t}, a_{t-25:t-1}]$ where $q_t$ and $\dot q_t$ denote joint positions and velocities, $w_t^{root}$ refers to the root angular velocity, $g_t$ is the gravity vector and $a_{t-1}$ is the previous action. These terms are stacked over the past 25 steps to form the proprioceptive input. The goal state is defined as $s_t^{g-deploy}\triangleq [\hat{h}_{t+5}, \hat{\theta}_{t+5}^{root}\ominus \theta_t^{root}, \hat{v}_{t+5}^{root}, \hat{w}_{t+5}^{root}, \hat{p}_{t+5} - \hat{p}_{t+5}^{root}]$, where $\hat{h}_{t+1}$ is the reference pose height, $\hat{\theta}_{t+1}^{root}$ and $\theta_t^{root}$ are the reference and current root orientations, $\hat{v}_{t+1}^{root}$ and $\hat{w}_{t+1}^{root}$ are the reference root linear and angular velocities, $w_t^{root}$ is the current root angular velocity, $\hat{p}_{t+1}$ and $\hat{p}_{t+1}^{root}$ are the reference rigid body and root positions. These terms are stacked over the future 5 steps. The first four terms are rotated into the current pose's local coordinate while the last term is rotated into the reference pose's local frame.

\begin{table*}[t]
    \centering
    \small
    \setlength{\tabcolsep}{1.5pt}
    \renewcommand{\arraystretch}{0.7}
    \begin{tabular}{l *{4}{c} *{3}{c}}
        \toprule
        \textbf{Methods} & \multicolumn{4}{c}{\textbf{All AMASS Train Dataset}} & \multicolumn{3}{c}{\textbf{Successful AMASS Train Dataset}} \\
         & SR↑ & MPKPE↓ & Vel-Dist↓ & Acc-Dist↓  & MPKPE↓ & Vel-Dist↓ & Acc-Dist↓ \\
        \midrule
        \rowcolor{grey!20}
        (a) Compare with Baselines & & & &  & & & \\
        \midrule
        OmniH2O~\cite{he2024omnih2ouniversaldexteroushumantohumanoid} &$84.58{\text{\tiny$\pm0.386$}}$ & $88.91{\text{\tiny$\pm0.067$}}$ & $7.09{\text{\tiny$\pm0.008$}}$& $3.98{\text{\tiny$\pm0.002$}}$& $88.14{\text{\tiny$\pm0.291$}}$ & $7.08{\text{\tiny$\pm0.012$}}$ & $3.82{\text{\tiny$\pm0.006$}}$ \\
        Exbody2~\cite{ji2025exbody2advancedexpressivehumanoid} & $86.39{\text{\tiny$\pm0.143$}}$& $86.77{\text{\tiny$\pm0.199$}}$ & $6.10{\text{\tiny$\pm0.071$}}$ & $3.63{\text{\tiny$\pm0.001$}}$ & $86.25{\text{\tiny$\pm0.045$}}$& $6.03{\text{\tiny$\pm0.001$}}$& $3.48{\text{\tiny$\pm0.006$}}$\\
        Dagger without CVAE & $88.03{\text{\tiny$\pm0.221$}}$ &$85.01{\text{\tiny$\pm0.185$}}$  & $5.63{\text{\tiny$\pm0.003$}}$ & $2.95{\text{\tiny$\pm0.001$}}$ & $84.78{\text{\tiny$\pm0.141$}}$ & $5.05{\text{\tiny$\pm0.003$}}$ & $2.40{\text{\tiny$\pm0.001$}}$ \\
        Train from Scratch & $72.22{\text{\tiny$\pm0.262$}}$ & $104.49{\text{\tiny$\pm0.453$}}$ & $8.79{\text{\tiny$\pm0.008$}}$ & $6.33{\text{\tiny$\pm0.005$}}$& $98.94{\text{\tiny$\pm0.396$}}$ & $8.44{\text{\tiny$\pm0.003$}}$ & $6.34{\text{\tiny$\pm0.003$}}$ \\
        Ours &\textbf{\boldmath $91.82{\text{\tiny$\pm0.114$}}$} & \textbf{\boldmath $82.57{\text{\tiny$\pm0.097$}}$} & \textbf{\boldmath$4.22{\text{\tiny$\pm0.001$}}$} & \textbf{\boldmath $1.79{\text{\tiny$\pm0.001$}}$}  & \textbf{\boldmath $78.95{\text{\tiny$\pm0.121$}}$} & \textbf{\boldmath $3.67{\text{\tiny$\pm0.001$}}$} & \textbf{\boldmath $1.39{\text{\tiny$\pm0.001$}}$} \\
        \midrule
        \rowcolor{grey!20}
        (b) Ablation with Architecture Design  & & & & & & & \\
        \midrule
        Actor with Explicit Reference & $88.20{\text{\tiny$\pm0.137$}}$ & $85.33{\text{\tiny$\pm0.068$}}$ & $5.76{\text{\tiny$\pm0.005$}}$ & $3.03{\text{\tiny$\pm0.001$}}$ & $84.85{\text{\tiny$\pm0.042$}}$ & $4.57{\text{\tiny$\pm0.002$}}$ & $2.15{\text{\tiny$\pm0.001$}}$ \\
        Actor without Explicit Reference &\textbf{\boldmath $91.82{\text{\tiny$\pm0.114$}}$} & \textbf{\boldmath $82.57{\text{\tiny$\pm0.097$}}$} & \textbf{\boldmath$4.22{\text{\tiny$\pm0.001$}}$} & \textbf{\boldmath $1.79{\text{\tiny$\pm0.001$}}$}  & \textbf{\boldmath $78.95{\text{\tiny$\pm0.121$}}$} & \textbf{\boldmath $3.67{\text{\tiny$\pm0.001$}}$} & \textbf{\boldmath $1.39{\text{\tiny$\pm0.001$}}$}\\
        \midrule
        \rowcolor{grey!20}
        (c) Ablation with KL Residual & & & & & & & \\
        \midrule
        KL without Residual &$85.39{\text{\tiny$\pm0.085$}}$ &$91.35{\text{\tiny$\pm0.163$}}$ &$6.01{\text{\tiny$\pm0.003$}}$ &$4.68{\text{\tiny$\pm0.004$}}$ &$90.84{\text{\tiny$\pm0.093$}}$ &$6.01{\text{\tiny$\pm0.001$}}$ &$4.23{\text{\tiny$\pm0.001$}}$ \\
        KL with Residual &\textbf{\boldmath $91.82{\text{\tiny$\pm0.114$}}$} & \textbf{\boldmath $82.57{\text{\tiny$\pm0.097$}}$} & \textbf{\boldmath$4.22{\text{\tiny$\pm0.001$}}$} & \textbf{\boldmath $1.79{\text{\tiny$\pm0.001$}}$}  & \textbf{\boldmath $78.95{\text{\tiny$\pm0.121$}}$} & \textbf{\boldmath $3.67{\text{\tiny$\pm0.001$}}$} & \textbf{\boldmath $1.39{\text{\tiny$\pm0.001$}}$} \\
        \midrule
        \rowcolor{grey!20}
        (d) Ablation with KL Coefficient & & & & & & & \\
        \midrule
        KL Coef = 1.0 & $81.39{\text{\tiny$\pm0.038$}}$ & $97.12{\text{\tiny$\pm0.216$}}$ & $7.78{\text{\tiny$\pm0.008$}}$ & $6.02{\text{\tiny$\pm0.006$}}$ & $96.41{\text{\tiny$\pm0.153$}}$ & $5.40{\text{\tiny$\pm0.001$}}$ & $2.41{\text{\tiny$\pm0.001$}}$ \\
        KL Coef = 0.1 &\textbf{\boldmath $91.82{\text{\tiny$\pm0.114$}}$} & \textbf{\boldmath $82.57{\text{\tiny$\pm0.097$}}$} & \textbf{\boldmath$4.22{\text{\tiny$\pm0.001$}}$} & \textbf{\boldmath $1.79{\text{\tiny$\pm0.001$}}$}  & \textbf{\boldmath $78.95{\text{\tiny$\pm0.121$}}$} & \textbf{\boldmath $3.67{\text{\tiny$\pm0.001$}}$} & \textbf{\boldmath $1.39{\text{\tiny$\pm0.001$}}$}\\
        KL Coef = 0.01 & $86.13{\text{\tiny$\pm0.012$}}$  &$89.77{\text{\tiny$\pm0.034$}}$  &$7.99{\text{\tiny$\pm0.005$}}$ & $5.47{\text{\tiny$\pm0.002$}}$&  $88.21{\text{\tiny$\pm0.037$}}$  & $5.23{\text{\tiny$\pm0.003$}}$ & $2.37{\text{\tiny$\pm0.001$}}$ \\
        KL Coef = 0.001 & $80.63{\text{\tiny$\pm0.101$}}$& $105.49{\text{\tiny$\pm0.265$}}$& $9.13{\text{\tiny$\pm0.009$}}$ & $7.98{\text{\tiny$\pm0.002$}}$ & $104.52{\text{\tiny$\pm0.198$}}$  &$5.38{\text{\tiny$\pm0.001$}}$ & $2.40{\text{\tiny$\pm0.001$}}$\\
        \midrule
        \rowcolor{grey!20}
        (e) Ablation with Future Window Size & & & & &  & &\\
        \midrule
        Window Size = 1 & $90.62{\text{\tiny$\pm0.103$}}$ & $86.02{\text{\tiny$\pm0.077$}}$ & $5.21{\text{\tiny$\pm0.003$}}$ & $3.12{\text{\tiny$\pm0.003$}}$& $85.06{\text{\tiny$\pm0.096$}}$&  $4.77{\text{\tiny$\pm0.001$}}$ & $2.91{\text{\tiny$\pm0.001$}}$ \\
        Window Size = 5 &\textbf{\boldmath $91.82{\text{\tiny$\pm0.114$}}$} &  $82.57{\text{\tiny$\pm0.097$}}$ & \textbf{\boldmath$4.22{\text{\tiny$\pm0.001$}}$} & \textbf{\boldmath $1.79{\text{\tiny$\pm0.001$}}$}  & $78.95{\text{\tiny$\pm0.121$}}$ & \textbf{\boldmath $3.67{\text{\tiny$\pm0.001$}}$} & \textbf{\boldmath $1.39{\text{\tiny$\pm0.001$}}$}\\
        Window Size = 10 & $91.25{\text{\tiny$\pm0.015$}}$ & \textbf{\boldmath $82.19{\text{\tiny$\pm0.138$}}$} &$4.76{\text{\tiny$\pm0.005$}}$ &$1.95{\text{\tiny$\pm0.001$}}$ & \textbf{\boldmath $78.72{\text{\tiny$\pm0.007$}}$}& $3.93{\text{\tiny$\pm0.002$}}$ &$1.88{\text{\tiny$\pm0.001$}}$  \\
        Window Size = 20 & $90.31{\text{\tiny$\pm0.111$}}$ & $82.93{\text{\tiny$\pm0.098$}}$& $5.01{\text{\tiny$\pm0.006$}}$& $1.78{\text{\tiny$\pm0.001$}}$& $80.28{\text{\tiny$\pm0.142$}}$& $4.31{\text{\tiny$\pm0.001$}}$& $1.48{\text{\tiny$\pm0.001$}}$\\
        \midrule
        \rowcolor{grey!20}
        (f) Ablation with Latent Dimension & & & & & & &\\
        \midrule
        Latent Dimension = 32 & $88.89{\text{\tiny$\pm0.053$}}$ & $87.24{\text{\tiny$\pm0.172$}}$ &$4.79{\text{\tiny$\pm0.004$}}$ &$2.36{\text{\tiny$\pm0.003$}}$ &$86.58{\text{\tiny$\pm0.079$}}$ &$4.65{\text{\tiny$\pm0.001$}}$ & $2.21{\text{\tiny$\pm0.001$}}$  \\
        Latent Dimension = 64 &\textbf{\boldmath $91.82{\text{\tiny$\pm0.114$}}$} & \textbf{\boldmath $82.57{\text{\tiny$\pm0.097$}}$} & $4.22{\text{\tiny$\pm0.001$}}$ & \textbf{\boldmath $1.79{\text{\tiny$\pm0.001$}}$}  & \textbf{\boldmath $78.95{\text{\tiny$\pm0.121$}}$} & \textbf{\boldmath $3.67{\text{\tiny$\pm0.001$}}$} & \textbf{\boldmath $1.39{\text{\tiny$\pm0.001$}}$} \\
        Latent Dimension = 128 &  $90.43{\text{\tiny$\pm0.132$}}$& $83.75{\text{\tiny$\pm0.059$}}$ & $4.46{\text{\tiny$\pm0.002$}}$&$2.14{\text{\tiny$\pm0.002$}}$ &$82.92{\text{\tiny$\pm0.129$}}$  &$4.43{\text{\tiny$\pm0.004$}}$   & $2.08{\text{\tiny$\pm0.001$}}$ \\
        Latent Dimension = 256 &$90.71{\text{\tiny$\pm0.098$}}$ &$84.18{\text{\tiny$\pm0.014$}}$ &\textbf{\boldmath $4.21{\text{\tiny$\pm0.002$}}$} &$1.97{\text{\tiny$\pm0.002$}}$ &$83.74{\text{\tiny$\pm0.031$}}$ & $4.19{\text{\tiny$\pm0.001$}}$ &$1.73{\text{\tiny$\pm0.001$}}$ \\
    \bottomrule
    \end{tabular}
    
    \caption{\textbf{Baselines Comparison and Ablation Studies} We compare our method with four baseline approaches and observe that it consistently outperforms them. Additionally, we conduct ablation studies on architectural design, KL residuals, KL coefficient, future window size, and latent dimensionality. The final configuration is selected based on the best overall performance observed across these settings.}
    \vspace{-5mm}
    \label{table4}
\end{table*}
\noindent \textbf{Modeling Diversity with CVAE}
Our target is to obtain a deployable policy $\pi^{deploy}(a_t|s_t^{p-deploy}, s_t^{g-deploy})$ which maintains the expressiveness of the oracle policy to the maximum. Given the deployable state space is partial and may lead to motion ambiguity, We model the deployable policy as a conditional variational autoencoder (CVAE) to model the diversity. Specifically, we have a variational encoder $\varepsilon(z_t|s_t^{p-orcale}, s_t^{g-oracle})$ that computes the latent code distribution based on the same observation with the teacher policy, a decoder $D(a_t|s_{t}^{p-deploy},z_t)$ that produces action. We employ a learned conditional prior $\rho(z_t|s_t^{p-deploy}, s_t^{g-deploy})$ following prior works which allows the model to learn different distributions based on proprioception. Using the evidence lower bound, we have the objective function as:

\begin{equation}
\begin{aligned}
    \log &P(a_t|s_t^{p-d}, s_t^{g-d})\ge E_{\varepsilon(z|s_t^{p-o}, s_t^{g-o})} [\log  D(a_t|s_{t}^{p-d},z_t)\\& - D_{KL}(\varepsilon(z_t|s_t^{p-o}, s_t^{g-o})||\rho(z_t|s_t^{p-d}, s_t^{g-d}))]
\end{aligned}
\end{equation}
where the prior, encoder and decoder are all modeled as diagonal Gaussian distribution. Following prior works, we model the encoder as a residual to the prior and have:
\begin{equation}
    \begin{aligned}
    \rho(z_t|s_t^{p-d}, s_t^{g-d})=\textit{N} &(\mu^\rho(s_t^{p-d}, s_t^{g-d}), \sigma^\rho(s_t^{p-d}, s_t^{g-d})) \\
    \varepsilon(z_t|s_t^{p-o},s_t^{g-o})=\textit{N} &(\mu^\rho(s_t^{p-d},s_t^{g-d})+\mu^{\varepsilon}(s_t^{p-o}, s_t^{g-o}), \\ &\sigma^{\varepsilon}(s_t^{p-o}, s_t^{g-o}))\\ D(a_t|s_t^{p-d},z_t)= \textit{N}&(\mu^{D}(s_t^{p-d}, z_t), \sigma^{D}(s_t^{p-d},z_t)) 
\end{aligned}
\end{equation}

To optimize the loss function, we use the action supervision signal provided by the oracle policy and learn the deployable policy in an online distillation fashion. The loss function derived from the objective above can be formulated as:
\begin{equation}
L = L_{action} + \beta L_{KL}
\end{equation}
where $L_{action}=||a_t^{deploy}-a_t^{oracle}||_2^2$ and $L_{KL}$ is the KL divergence between the prior and encoder.
\subsection{Fast Adaption on top of a Universal Policy}
\label{sec:E}
In this stage, we focus on motion references that the universal policy fails to reliably track. These motions typically exhibit two challenging characteristics: they lie far outside the training data distribution, and they involve highly dynamic behaviors that approach the physical limits of the humanoid robot.

To address these cases, we introduce a lightweight residual decoder $D(\widetilde{a_t}|{s_t}^{p-deploy}, {s_t}^{g-deploy})$ appended to the universal motion tracking framework, to enable fast adaptation to challenging motions. During training, we initialize the model with the universal policy checkpoint and freeze all existing parameters. Only the residual decoder is fine-tuned using reinforcement learning, the reward is the same as Section~\ref{sec:C}. The input to the residual decoder is identical to that of the prior network. We directly use explicit motion references instead of latent representations for two main reasons. First, such out-of-distribution motions may not be adequately captured by the learned CVAE prior. Second, the goal at this stage is to quickly overfit to specific motion sequences rather than to maintain generalization. The final action applied to the environment is obtained by summing the output of the universal policy and the residual correction:
\begin{equation}
    a_t^{final} = a_t + \widetilde{a_t},
\end{equation}
where $a_t$ is the action predicted by the frozen universal policy, $\widetilde{a_t}$ is the residual generated by the residual decoder.




\section{Experimental Results}
\subsection{Experiment Setup}
We evaluate UniTracker in both simulated and real-world environments. In simulation, policies are trained using IsaacGym~\citep{makoviychuk2021isaacgymhighperformance} with 8192 parallel environments under domain randomization~\cite{tobin2017domainrandomizationtransferringdeep}. The training data is derived from the AMASS dataset~\cite{mahmood2019amassarchivemotioncapture} and filtered by PHC~\citep{luo2023perpetualhumanoidcontrolrealtime}. We further validate our approach through sim-to-sim transfer by deploying the trained policies in MuJoCo, and conduct extensive ablation and comparative studies. Performance is assessed using four key metrics: Success Rate (SR), Mean Per Keypoint Position Error (MPKPE), Velocity Distance(Vel-Dist) and Acceleration Distance(Acc-Dist). SR reflects the overall viability and stability of the policy; MPKPE quantify the accuracy of keypoint tracking in the world coordinate frame; Vel-Dist quantifies the difference in joint velocities between the reference motion and the executed motion of the robot; Acc-Dist quantifies the difference in joint acceleration between the reference motion and the executed motion of the robot. For real-world evaluation, we deploy UniTracker on the Unitree G1 humanoid robot, which stands 1.3 meters tall and has 29 degrees of freedom. We control 23 DoFs by locking the 6 wrist joints.

\subsection{Baselines}
\textbf{What kind of motion tracker yields the best tracking performance?} We include two internal baselines to assess the effectiveness of each component in our framework:
(1) a universal policy trained from scratch without leveraging the teacher-student architecture;
(2) a teacher-student framework using DAgger~\cite{ross2011reductionimitationlearningstructured} in the second stage, but without incorporating CVAE-based modeling.
In addition, we compare UniTracker against recent state-of-the-art approaches for humanoid motion tracking, including OmniH2O\cite{he2024omnih2ouniversaldexteroushumantohumanoid} and Exbody2\cite{ji2025exbody2advancedexpressivehumanoid}. Experimental results presented in Table~\ref{table4} demonstrate that UniTracker consistently outperforms all baselines and prior methods across all evaluation metrics. 

\subsection{Ablation Studies}

\textbf{What model architecture best balances diversity, expressiveness and robustness?} To answer this question, we first investigated whether the RL actor (the decoder in the CVAE) should explicitly take the reference motion as part of its input. We evaluated two variants: one where the actor input is $({s_t}^{p-deploy}, {s_t}^{g-deploy}, z_t)$ and another where the reference motion is excluded. The results show that when the actor receives the reference motion directly, the influence of the latent variable $z$ vanishes. In this case, the behavior of the model closely resembles that of a standard DAgger setup, as the strong reference input causes the policy to ignore the latent guidance. 

\begin{figure}
    \centering
    \vspace{-2mm}
    \includegraphics[width=\columnwidth]{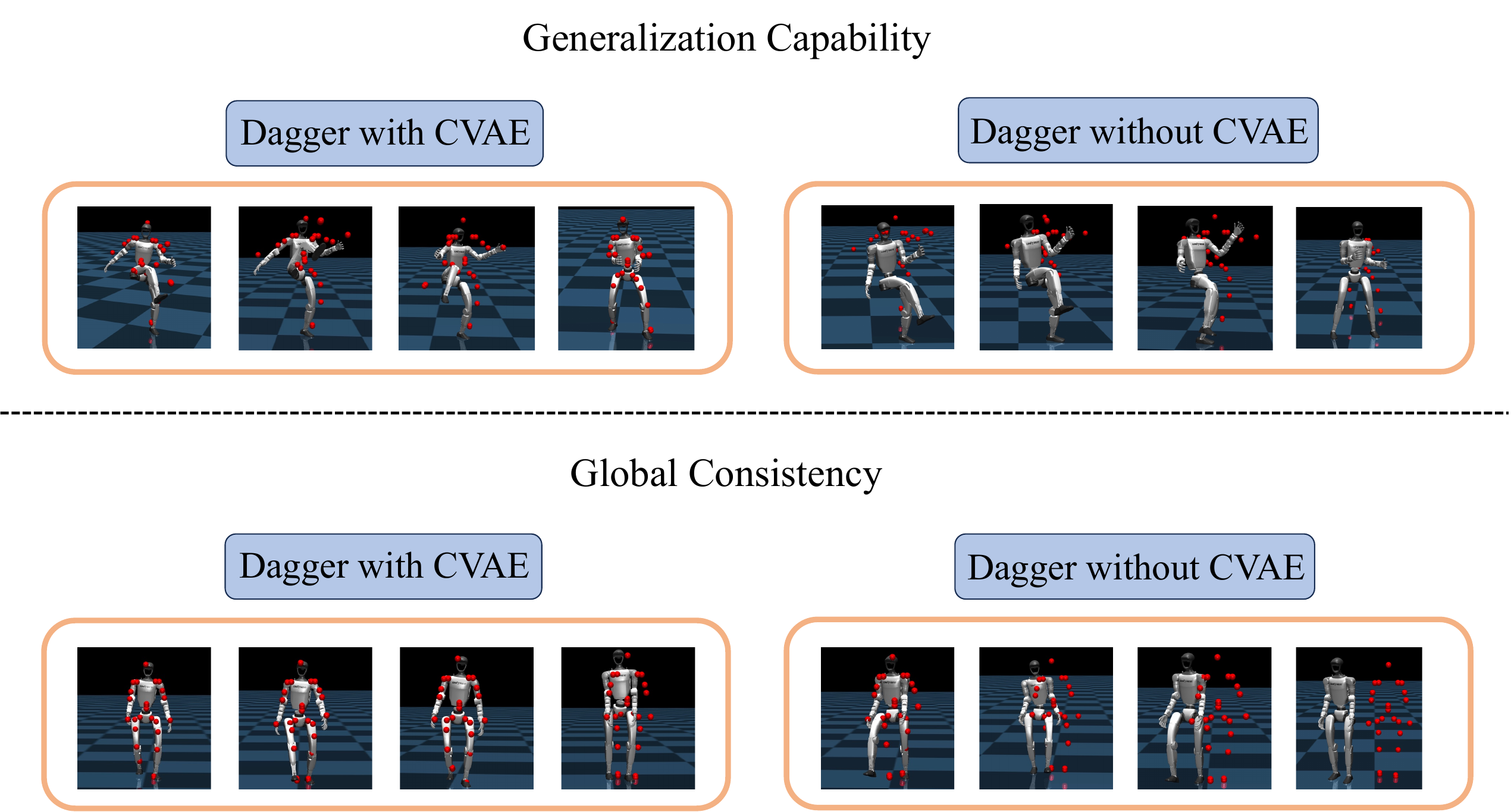}
    \caption{Generalization Ability and Global Consistency of UniTracker in the MuJoCo Simulator}
    \vspace{-5mm}
    \label{fig:gloabl_info}
\end{figure}

In Figure~\ref{fig:gloabl_info}, we evaluate the generalization capability of UniTracker by test the policy in MuJoCo on a challenging crescent kick motion that is unseen during training. We observe that UniTracker successfully tracks this previously unseen behavior, while the variant without CVAE modeling fails to reproduce the motion effectively. This result highlights the importance of structured latent motion modeling for generalization to out-of-distribution motions.
Additional real-world out-of-distribution (OOD) examples are provided in the supplementary video, further demonstrating the robustness and adaptability of our approach under novel motion inputs.

In Table~\ref{tab:robust-test}, we evaluate the robustness of our policy by incrementally adding noise to the observations. In this table, a noise level of 0 indicates no observation noise, while higher levels correspond to increasing noise intensity. We observe that UniTracker consistently outperforms the variant without CVAE in both Success Rate (SR) and Mean Per Keypoint Position Error (MPkPE). Moreover, UniTracker exhibits a slower degradation in performance as noise increases, indicating stronger robustness under observation perturbations. 

In Figure~\ref{fig:gloabl_info}, we evaluate the global awareness of our CVAE-based framework using a walking sequence in the MuJoCo simulator. The results show that, under our framework, the robot is able to walk in a straight line while closely following the reference trajectory. In contrast, the framework without CVAE gradually deviates from the reference path, demonstrating the superiority of our approach in maintaining global consistency.

\begin{figure}
    \centering
    \vspace{-2mm}
    \includegraphics[width=\columnwidth]{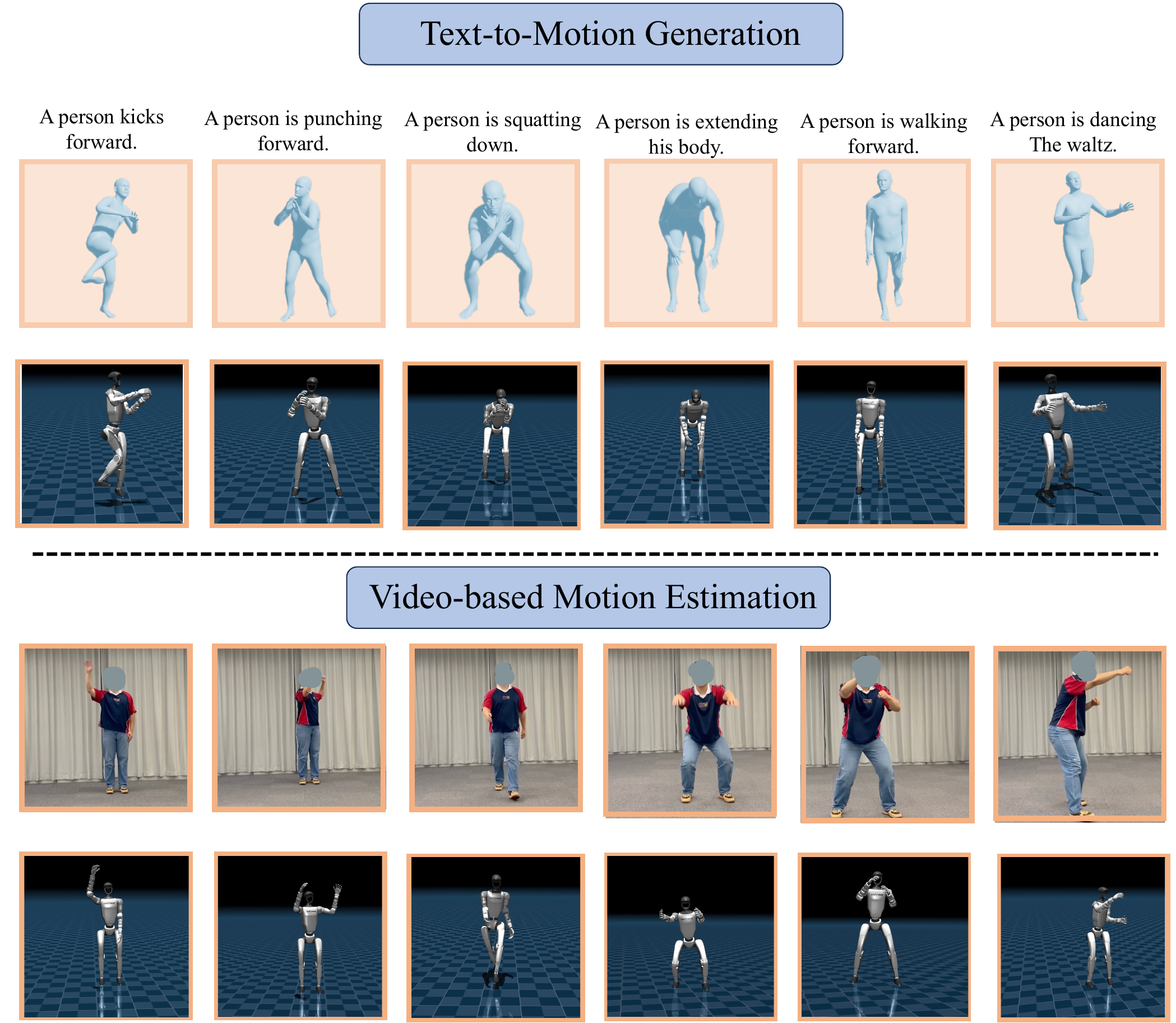}
    \caption{\textbf{The Outcome of Downstream Applications in mujoco:} We evaluate text-to-motion generation and video-based motion estimation in the muJoCo simulator.}
    \vspace{-6mm}
    \label{fig:application}
\end{figure}

\textbf{What hyperparameter choices lead to the best performance in whole-body motion tracking?}
We conduct ablation studies on three key hyperparameters in our framework: the CVAE latent dimension, the future window size of the reference motion used in the CVAE prior, and the weight of the KL loss term.
For the latent dimension, we experiment with values of 32, 64, 128, and 256. We observe that a dimension of 64 yields the best performance. Larger latent sizes make the CVAE harder to train and do not lead to further improvements. Results are shown in Table~\ref{table4}
For the future window, which controls how many future frames of reference motion are included in the prior input, we test window sizes of 1, 5, 10, and 20 frames. The results indicate that using 5 future frames provides the best trade-off between responsiveness and stability.
Lastly, we tune the KL loss coefficient with values of 1.0, 0.1, 0.01, and 0.001. We find that setting the weight to 0.1 leads to the best policy performance, effectively balancing latent regularization and reconstruction quality. 
\vspace{-2mm}
\subsection{Downstream Applications}
\vspace{-2mm}
We evaluate two downstream applications of our framework: motion generation and video-based motion estimation. For motion generation, we use the MDM~\cite{tevet2022humanmotiondiffusionmodel} model to produce SMPL-based~\cite{SMPL:2015} motion sequences conditioned on text input. These motions are then retargeted to the G1 humanoid robot for tracking. Both MuJoCo~\cite{todorov2012mujoco} simulation and real-world results demonstrate that our policy can accurately track motions generated from text prompts.
For video estimation, we record a motion sequence using a monocular camera, and convert it to SMPL format using GVHMR~\cite{shen2024gvhmr}. The resulting motion is retargeted to the G1 robot. Experiments in both MuJoCo and the real world show that our policy successfully tracks the reconstructed motions.
These two applications highlight the strong generalization capability of our policy across different types of reference inputs.

    

\subsection{Light Weight Fast Adaption}
\begin{figure}
    \centering

    \includegraphics[width=\columnwidth]{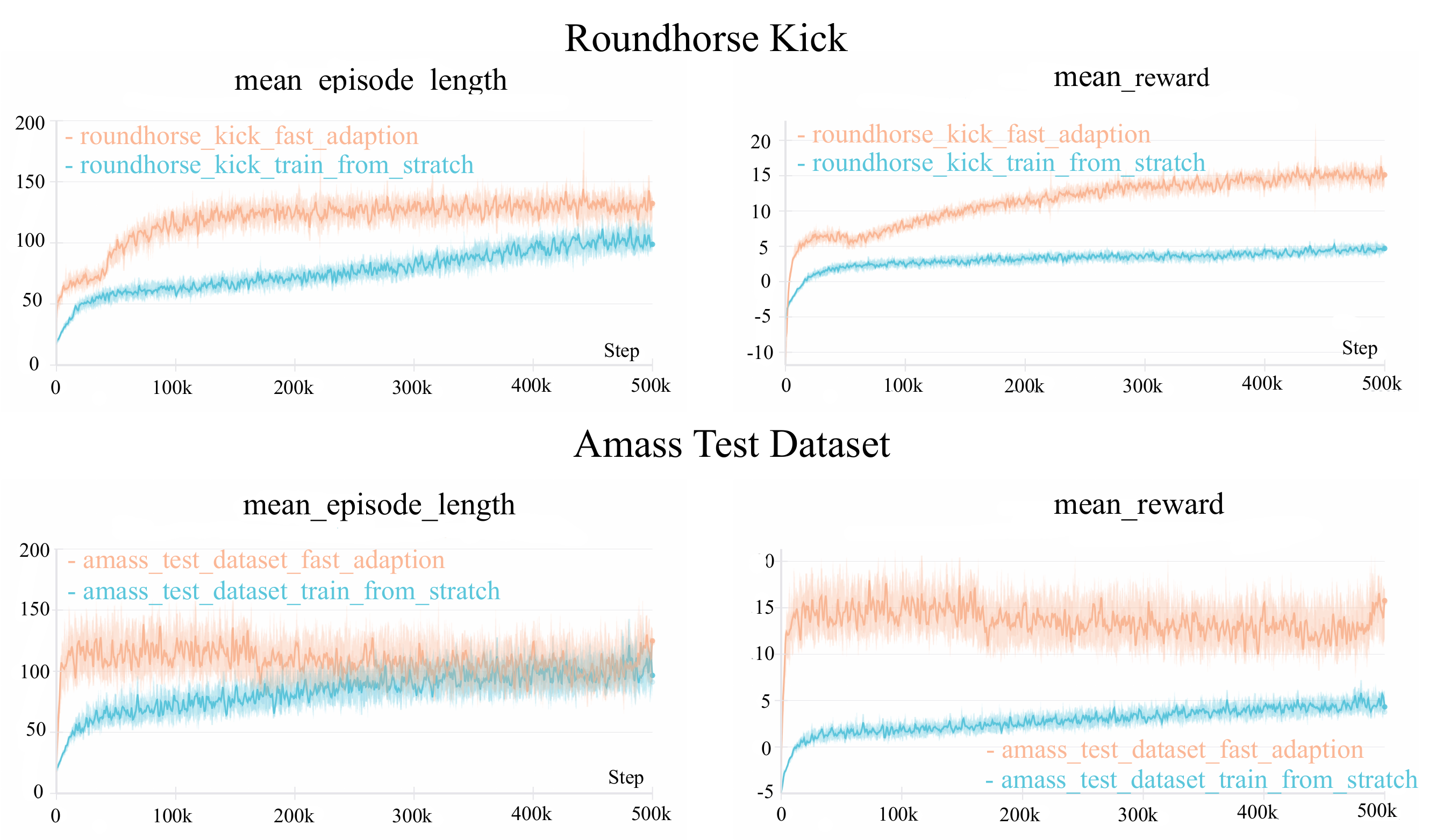}
    \caption{Fast Adaption of Challenging Motions}
    \vspace{-8mm}
    \label{fig:fast_adaption}
\end{figure}

We compared our fast adaptation module against a training-from-scratch baseline. In the latter, the observations of our model remain consistent with those of the second-stage student policy, and the reward function is identical to that used in the fast adaptation module. For evaluation, we selected one challenging motion sequence along with the entire AMASS test dataset. As shown in Figure~\ref{fig:fast_adaption}, our fast adaptation approach converges substantially faster than training from scratch in terms of both episode length and cumulative reward. These results underscore the advantages of our second-stage universal policy and highlight the effectiveness of the pretrained model in enabling rapid adaptation.

\begin{table}[t]
\centering
\scriptsize
\setlength{\tabcolsep}{2pt}
\renewcommand{\arraystretch}{0.5}
\begin{tabular}{l *{2}{>{\centering\arraybackslash}p{1.8cm}}}
  \toprule
  \textbf{Methods} & \multicolumn{2}{c}{\textbf{All AMASS Train Dataset}}\\
   & SR$\uparrow$ & MPKPE$\downarrow$ \\
  \midrule
  \rowcolor{gray!20}(a) Noise Level 0 & & \\
  \midrule
  Dagger without CVAE & $88.23{\text{\tiny$\pm0.105$}}$ & $84.75{\text{\tiny$\pm0.091$}}$\\
  \textbf{Ours}        & \textbf{\boldmath$91.81{\text{\tiny$\pm0.048$}}$} & \textbf{\boldmath$82.69{\text{\tiny$\pm0.116$}}$}\\
  \midrule
  \rowcolor{gray!20}(b) Noise Level 1 & & \\
  \midrule
  Dagger without CVAE & $85.78{\text{\tiny$\pm0.191$}}$ & $88.61{\text{\tiny$\pm0.312$}}$\\
  \textbf{Ours}        & \textbf{\boldmath$90.29{\text{\tiny$\pm0.094$}}$} & \textbf{\boldmath$83.83{\text{\tiny$\pm0.168$}}$}\\
  \midrule
  \rowcolor{gray!20}(c) Noise Level 2 & & \\
  \midrule
  Dagger without CVAE & $79.58{\text{\tiny$\pm0.066$}}$ & $93.79{\text{\tiny$\pm0.019$}}$\\
  \textbf{Ours}        & \textbf{\boldmath$86.78{\text{\tiny$\pm0.212$}}$} & \textbf{\boldmath$87.31{\text{\tiny$\pm0.182$}}$}\\
  \bottomrule
\end{tabular}
\captionof{table}{We evaluate the robustness of the policy by incrementally adding noise to the observations.}
\label{tab:robust-test} 
\end{table}

\section{Related Work}
\subsection{Whole-Body Controller for Humanoid Robots}
Whole-body control is essential for enabling humanoid robots to perform a wide variety of complex tasks. Prior to the rise of reinforcement learning, researchers primarily relied on traditional optimization-based control methods for humanoid whole-body control~\cite{1642100, chignoli2021mithumanoidrobotdesign, li2022dynamicwalkingbipedalrobots}. These approaches typically required explicit mathematical modeling of both the robot and its environment, followed by real-time optimization to compute the robot’s next action. However, such methods often struggle to adapt to environmental variations, resulting in limited robustness. Additionally, they impose heavy computational demands during online execution.

To overcome these limitations, reinforcement learning (RL) has emerged as a powerful alternative, offering the ability to learn adaptive, robust control policies directly from interaction with the environment without relying on explicit modeling. Current reinforcement learning–based whole-body controllers for humanoid robots can be categorized by the source of their control signals, including teleoperation~\cite{ze2025twistteleoperatedwholebodyimitation, lu2025mobiletelevisionpredictivemotionpriors, fu2024humanplushumanoidshadowingimitation,
he2024learninghumantohumanoidrealtimewholebody, qiu2025humanoidpolicyhuman, he2024omnih2ouniversaldexteroushumantohumanoid}, offline motion datasets~\cite{shao2025langwbclanguagedirectedhumanoidwholebody, shi2025adversariallocomotionmotionimitation,xie2025kungfubot, liao2025beyondmimicmotiontrackingversatile,  ji2025exbody2advancedexpressivehumanoid, cheng2024expressivewholebodycontrolhumanoid}, video-based motion estimation~\cite{mao2024learningmassivehumanvideos, allshire2025visualimitationenablescontextual}~\cite{qiu2025humanoidpolicyhuman, he2024omnih2ouniversaldexteroushumantohumanoid}, and high-level task commands~\cite{he2025hoverversatileneuralwholebody, xue2025unifiedgeneralhumanoidwholebody}. Teleoperation involves directly controlling the humanoid robot in real time using human input, often through motion capture systems or wearable sensors, allowing the robot to mimic human movements with high fidelity. Representative works in this area include Twist~\cite{ze2025twistteleoperatedwholebodyimitation} and H2O~\cite{he2024learninghumantohumanoidrealtimewholebody}, both of which adopt a two-stage teacher-student framework. The primary difference lies in the design of the policy's observation space. Offline motion datasets consist of pre-collected human or humanoid motion sequences, which are used as references for training control policies through motion imitation. Representative works are Exbody~\cite{cheng2024expressivewholebodycontrolhumanoid} and Exbody2~\cite{ji2025exbody2advancedexpressivehumanoid}, which begins by carefully curating an offline motion dataset and then decouples upperand lower-body motions as much as possible, aiming to maintain stability in the lower body while encourage diversity and expressiveness in the upper body. In addition, the most recent work, GMT~\cite{chen2025gmt}, is the first to demonstrate tracking of 8,000 motions using a single unified policy. Video-based motion estimation methods leverage visual input from videos to extract human motion data, which can then be used to guide humanoid robot control policies. This approach enables learning from large-scale, diverse motion sources without requiring direct human demonstration. A representative work is VideoMimic~\cite{mao2024learningmassivehumanvideos}, which develops a real-to-sim-to-real pipeline to model both the robot and its surrounding environment. Task commands refer to high-level, sparse control signals that specify desired outcomes or goals, such as walking direction or target position, rather than detailed joint-level motions, enabling efficient whole-body control through abstraction. Representative works include Hover~\cite{he2025hoverversatileneuralwholebody}and HugWBC~\cite{xue2025unifiedgeneralhumanoidwholebody}. Hover unifies multiple control modes into a single policy, enabling seamless transitions while retaining the strengths of each mode, thus providing a robust and scalable humanoid control solution. HugWBC designs a general task and behavior command space and employs techniques such as symmetrical loss and intervention training. This enables real-world humanoid robots to perform a variety of natural gaits—including walking, jumping, and hopping.

\section{Conclusion}
In this work, we present UniTracker, a unified and scalable framework for whole-body motion tracking in humanoid robots. Built upon a three-stage training pipeline, our approach begins with a privileged teacher policy that enables high-fidelity motion tracking and effective data curation. We then introduce a CVAE-based student policy that achieves robust deployment under partial observations by modeling motion diversity and implicitly incorporating global context. To further extend the system’s adaptability, we propose a lightweight residual decoder for fast adaptation to highly dynamic or out-of-distribution motions.
We validate UniTracker extensively in both simulation and real-world settings using a 29-DoF Unitree G1 humanoid. Our method successfully tracks over 8,100 motion sequences with a single policy, outperforming strong teacher-student baselines and prior methods in terms of accuracy, generalization, and robustness. The results demonstrate the effectiveness of combining generative modeling with hierarchical policy distillation and residual adaptation for expressive and general-purpose humanoid control.






\printbibliography
\end{document}